\documentclass[]{article}
\usepackage{proceed2e}

\usepackage{times}
\usepackage{amsmath}
\usepackage{amssymb}
\usepackage{amsthm}
\usepackage{listings}
\usepackage{algpseudocode}
\usepackage{algorithm}
\usepackage{graphicx}
\usepackage{subfigure}

\title{Structured Factored Inference: A Framework for Automated Reasoning in Probabilistic Programming Languages}

\author{ {\bf Avi Pfeffer} \\
Charles River Analytics \\
Cambridge, MA 02138 \\
apfeffer@cra.com \\
\And
{\bf Brian Ruttenberg} \\
Charles River Analytics \\
Cambridge, MA 02138 \\
bruttenberg@cra.com \\
\And
{\bf William Kretschmer}\thanks{ Work performed while interning at Charles River Analytics} \\
MIT \\
Cambridge, MA 02139 \\
kretsch@mit.edu
}

\begin{document}

\maketitle

\begin{abstract}
Reasoning on large and complex real--world models is a computationally difficult task, yet one that is required for effective use of many AI applications. A plethora of inference algorithms have been developed that work well on specific models or only on parts of general models. Consequently, a system that can intelligently apply these inference algorithms to different parts of a model for fast reasoning is highly desirable. We introduce a new framework called structured factored inference (SFI) that provides the foundation for such a system. Using models encoded in a probabilistic programming language, SFI provides a sound means to decompose a model into sub--models, apply an inference algorithm to each sub--model, and combine the resulting information to answer a query. Our results show that SFI is nearly as accurate as exact inference yet retains the benefits of approximate inference methods.
\end{abstract}

\section{INTRODUCTION}

Probabilistic modeling is at the core of many artificial intelligence (AI) applications. The complexity, richness, and diversity of these models are rapidly growing as AI takes on a larger role in everyday life. As a result, the efficiency of probabilistic inference is critical for effective use of these models. However, despite significant research into efficient algorithms and techniques, probabilistic inference remains a significant bottleneck in many real--world AI applications.

Many different algorithms have been explored to reason on general models. Unfortunately, no single algorithm performs sufficiently on every model, and often there are trade--offs that must be made. For example, sampling methods such as Metropolis-Hastings~\cite{chib1995understanding} are often the ``go to'' algorithms for reasoning on continuous models, but convergence can be painfully slow and they suffer from high variance estimates. Exact methods such as variable elimination~\cite{koller2009probabilistic} work well on discrete problems, but are intractable for all but the simplest models. Recent work on generalized variational inference~\cite{wainwright2008graphical} shows promise, but still requires some hand--tuning to work effectively. Once an algorithm has been found to work well on a specific problem, even slight modifications are no guarantee of continued success; adding a single continuous variable to an otherwise discrete problem can vastly affect the performance of the existing algorithm. Thus, AI developers are faced with selecting and configuring the appropriate algorithm that will work well on their problem, a task that is often more time consuming than constructing the actual model.

One solution to reduce this burden is to develop a method that can automatically select an algorithm that should perform well on one's specific problem. One major impediment to this approach is that the size and complexity of real--world models makes it difficult to determine the best single algorithm. Indeed, different algorithms might be appropriate for different parts of the model. For example, one algorithm might be appropriate for a continuous portion of the model while another is used for a discrete portion; indeed, the Rao--Blackwellization algorithm~\cite{doucet2000rao} exploits this fact. As a result, one approach to achieving automated inference is to not just select a single algorithm, but rather a \textit{set} of algorithms to apply to different parts of a model, and combine the results in the appropriate manner. Central to this approach is developing a sound method to decompose a model into manageable sub--models with the appropriate granularity. Such a method would provide a framework to intelligently select algorithms for different parts of a model, and the means to combine results to answer the query.

The emerging field of probabilistic programming (PP) provides the opportunity to support this decomposition framework. PP~\cite{koller1997effective,goodman2013principles} provides expressive and general purpose languages to encode a probabilistic model as an executable program. This allows one to leverage the power of programming languages to create rich and complex models, and use built--in inference algorithms that can operate on any model written in the language. More importantly, since the models are encoded as a program, we can use the program structure to understand the properties of a model before inference is even attempted. %From an automated inference perspective, PP can provide sound and proven techniques to decompose and analyze general probabilistic models.

We introduce a new PP--based inference framework called structured factored inference (SFI). SFI uses simple PP semantics to identify \textit{decomposition points} within a probabilistic model, creating an abstract hierarchy of sub--models. Each sub--model is independently reduced to a joint distribution over variables relevant to answering the query, using any inference method. Using factors to represent this joint distribution, the results are incorporated into the inference algorithms applied on other sub--models. The SFI framework brings many significant advantages. First, SFI provides the capability to apply \textit{decomposition strategies} to decomposition point so that sub--models can be created with small interfaces (and thus small joint distributions). For example, a strategy could choose to create a sub--model defined by a single decomposition point or combine several decomposition points into a single sub--model. Second, it has the ability to apply \textit{inference strategies} that choose algorithms to ``solve'' a particular sub--model during the inference process. This means that one need not decide on a single inference algorithm to apply to an entire model.

We show the benefits of SFI on three realistic models using a combination of exact and approximate inference algorithms. Our experiments show that even with simple strategies for decomposition and inference, probabilistic reasoning using SFI achieves performance equal to or better than approximate inference methods and is nearly as accurate as exact inference methods. The SFI framework is extremely general and expandable, providing the opportunity to use more complex decomposition strategies or intelligent inference strategies. The SFI framework has the potential to be the foundation for a general automated inference system.

\section{RELATED WORK}

Automated algorithm selection has been a long desired goal in computer science, with possibly the first formulation by Rice~\cite{rice1976algorithm}. As such, it has been applied to a variety of disciplines in the field, such as scientific computing~\cite{houstis2000pythia}, game theory~\cite{guerri2004learning}, and artificial intelligence problems such as satisfiability~\cite{xu2008satzilla}. Most efforts, however, focus on methods to analyze and learn how to apply the single best algorithm to solve a problem. For example, Guo~\cite{guo2003bayesian} uses Bayesian networks to learn and select the best algorithm to solve a problem; neither the problems or algorithms are specific, and can be applied generally to a variety of problems. Our SFI framework is complementary to much of this existing work. SFI can decompose complex models into smaller sub--models that these sophisticated learning algorithms can operate on, potentially providing even better performance than learning on a single model.

Probabilistic inference is unique in some respects as the independence properties of models provides the opportunity to apply many algorithms to different parts of a problem. There has not been significant amount of progress in the probabilistic modeling community, however, to address this or take advantage of it. Our approach is similar in spirit to the current work on black box variational inference~\cite{ranganath2013black}. These recent methods have attempted to reduce the programmer burden of configuring and applying variational inference to general probabilistic models, and in a sense are attempting to automatically find the best configuration of the algorithm that produces optimal inference results. While these approaches are promising, they still only consider a single algorithm. The decomposition strategies in SFI also bears similarity to structured variable elimination (SVE)~\cite{koller1997object}. Like SVE, SFI enjoys the benefits of decomposition in exploiting small interfaces and reusing work. However, SVE applies the same algorithm to each problem, whereas SFI is a general framework for decomposing problems and optimizing each sub--model separately. Finally, our work is similar to decomposition methods that solve maximum a posteriori (MAP) queries presented by Duchi et. al.~\cite{elidan2007using}. This work, however, only applies to specific decompositions of Markov random fields and only applies to MAP problems.

%SFI is a factored framework, which provides several advantages. First, the results of inference on a sub--model need to be captured and communicated to other parts of a model. From an engineering perspective, factors are a principled way to accomplish this. Second, many types of algorithms can be integrated into SFI, as factors do not preclude the use of sampling or other non--factor algorithms; such algorithms need only ingest and output factors, and the internal workings of the algorithm are not important to the framework. 

\section{PROBABILISTIC PROGRAMMING LANGUAGE}
\setlength{\textfloatsep}{6pt}
The fundamental concepts in SFI are strongly tied to probabilistic programming languages (PPL), and SFI has been implemented in a publicly available PPL. As such, understanding PPL semantics is critical for understanding SFI. However, since the focus of this paper is introducing the SFI concept, we present a simplified and abstract PPL for purposes of explanation. We call this abstract language SimplePPL. 

\subsection{SimplePPL Language}
The central concept in SimplePPL is a random variable (RV). Intuitively, a RV represents a random process that stochastically produces a value. For simplicity, we use an untyped language, so an RV can produce any value in the value set $T$, where $T$ is a countable finite set. A program $Q$ has a set of \textit{free} RVs $\mathcal{F}_Q$, and consists of a sequence of definitions of the form $RV = expr$.
The set of RVs defined by $Q$ is denoted $\mathcal{RV}_Q$. An RV is \textit{available} if it is either in $\mathcal{F}_Q$ or defined previously in $Q$. The set of available RVs with respect to an RV $r$ is denoted $\mathcal{A}_r$.

An expression defining an RV $r$ is one of the following:
\begin{itemize}
\vspace{-1mm}
\itemsep-0.5em
\item A value $v$
\item A primitive defining a probability distribution over values. Examples include $Flip(p)$, which produces true with probability $p$, and $Uniform(x, y)$.
\item $Apply(r_1, \dots, r_n, f)$, where $r_1, \dots ,r_n$ are available RVs and $f$ is a function $T^n \rightarrow T$. 
\item $Chain(r_1, f)$, where $r_1$ is an available RV and $f$ is a function $T \rightarrow \mathcal{Q}$, where $\mathcal{Q}$ is the space of programs such that for each $Q' \in \mathcal{Q}$, $\mathcal{F}_{Q'} \subseteq \mathcal{A}_r$ and the final RV in the program is named ``outcome''.
\end{itemize}

\subsection{SimplePPL Semantics}
Although there are clear semantics for recursive programs in SimplePPL, for simplicity in this paper it will suffice to assume that the expansions of a program $Q$ are non--recursive and thus finite. Under the SimplePPL semantics, each program $Q$ defines a conditional probability distribution $P(\mathcal{RV}_Q | \mathcal{F}_Q)$. This is achieved by defining, for each RV $r$ defined in $Q$, a conditional distribution $P(r | \mathcal{A}_r)$ and then using the chain rule so that
\begin{equation*}
P(\mathcal{RV}_Q | \mathcal{F}_Q) = \prod_{r \in \mathcal{RV}_Q} P(r | \mathcal{A}_r)
\end{equation*}
$P(r | \mathcal{A}_r)$ is defined as follows:
\begin{itemize}
\vspace{-1mm}
\itemsep-0.5em
\item For $r$ defined by value $v$, $P(r | \mathcal{A}_r)$ assigns probability $1$ to $v$.

\item For $r$ defined by a primitive distribution $\pi$, $P(r | \mathcal{A}_r)$ is $P(\pi)$.

\item For $r$ defined by $Apply(r_1, \dots, r_n, f)$, by assumption $\lbrace r_1, \dots, r_n \rbrace \subseteq \mathcal{A}_r$. Therefore, $P(r | \mathcal{A}_r)$ assigns probability $1$ to $f(v_1, \dots, v_n)$ for any values $v_1, \dots, v_n$ in the support of $r_1, \dots, r_n$.

\item For $r$ defined by $Chain(r_1, f)$, let $v_1$ be the value of $r_1$. Let $Q'$ be the program $f(v_1)$. By induction, $Q'$ defines a distribution $P_{v_1}(\mathcal{RV}_{Q'} | \mathcal{F}_{Q'})$. By definition, $\mathcal{F}_{Q'} = \mathcal{A}_r$ and because ``outcome'' $\in \mathcal{RV}_{Q'}$, $Q'$ defines a distribution $P_{v_1}(outcome | \mathcal{A}_r)$. As $r_1 \in \mathcal{A}_r$, the distribution $P(r | \mathcal{A}_r)$ is equal to $P_{v_1}(outcome | \mathcal{A}_r)$ for any value $v_1 \in r_1$.
\end{itemize}

SimplePPL is purely functional, and evidence (conditions or constraints) can be applied as part of a query on a model. While SimplePPL lacks the complexity of many PPLs, it is as expressive as full--fledged PPLs.

\section{STRUCTURED FACTORED INFERENCE}

\begin{figure}[t]
\begin{lstlisting}[basicstyle=\scriptsize]
a = Flip(0.6)
b = Chain(a, f)
c = Chain(a, f)

f(true) = {
  x1 = Flip(0.9)
  y1 = Flip(0.8)
  z1 = Apply(x1, y1, (b1, b2) => b1&&b2)
  x2 = Flip(0.7)
  y2 = Flip(0.8)
  z2 = Apply(x2, y2, (b1, b2) => b1&&b2)
  outcome = Apply(z1, z2, (b1, b2) => b1||b2)
}
f(false) = {
  x1 = Flip(0.1)
  y1 = Flip(0.8)
  z1 = Apply(x1, y1, (b1, b2) => b1&&b2)
  x2 = Flip(0.2)
  y2 = Flip(0.8)
  z2 = Apply(x2, y2, (b1, b2) => b1&&b2)
  outcome = Apply(z1, z2, (b1, b2) => b1||b2)
}
\end{lstlisting}
\vspace{-1mm}
\caption{SimplePPL code for a small model. The ``\texttt{=>}'' denotes an anonymous function in SimplePPL.\label{exampleModel}}
%\vspace{-3mm}
\end{figure}

\begin{figure}[t]
\centering % was .9 columnwidth
\includegraphics[width=0.80\columnwidth]{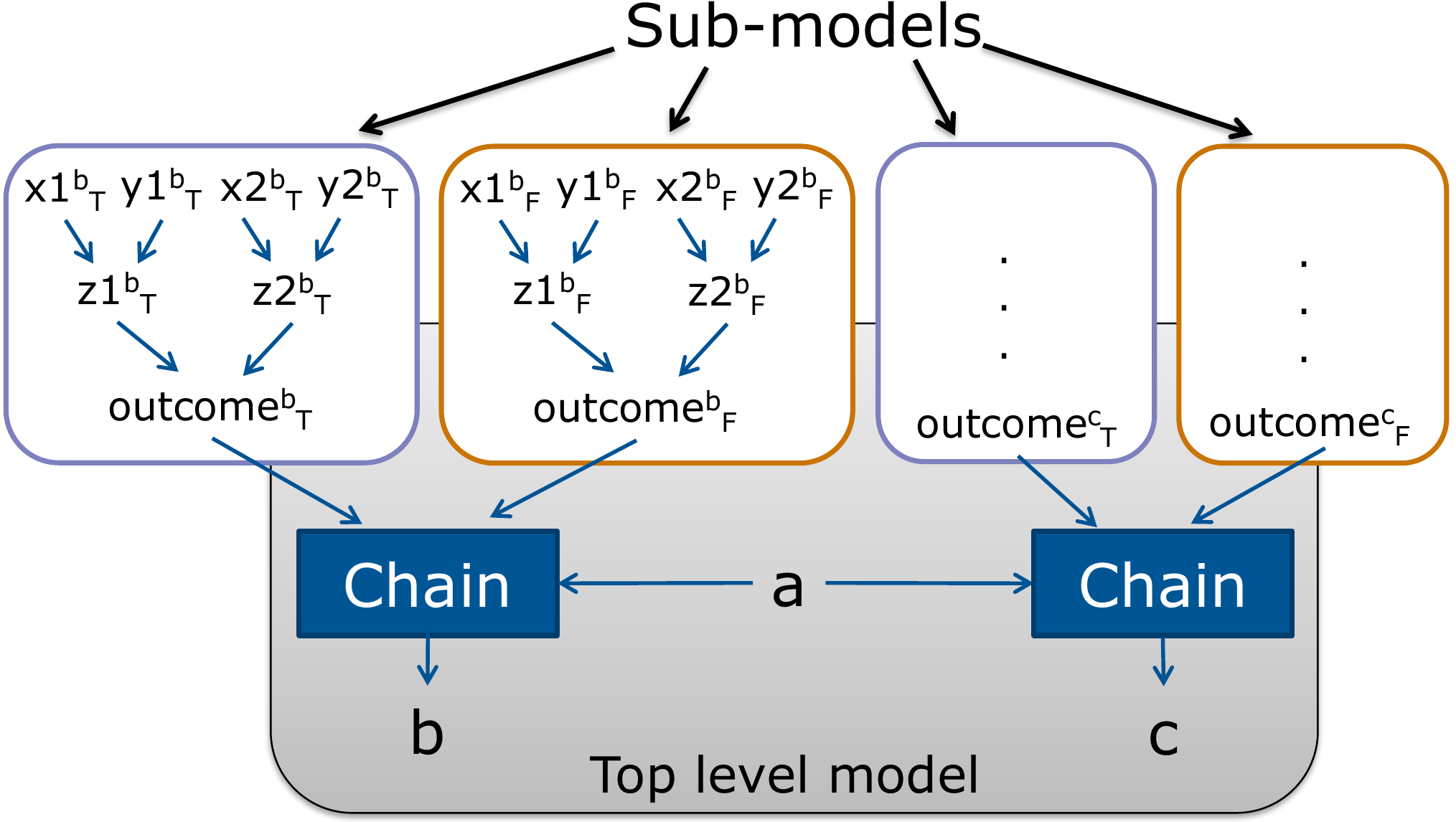}
\vspace{-1mm}
\caption{The model presented in Fig.~\ref{exampleModel} \label{LowLevel}}
\vspace{1mm}
\end{figure}

\subsection{Overview}

There is a simple intuition behind SFI: If a model can be broken down into smaller sub--models (i.e., programs) that can solved independently (i.e., marginalizing out non--relevant variables), then different algorithms can be applied to different parts of a model. Combined with methods for intelligent inference algorithm selection, this framework could then lead to improved inference on a wide variety of problems. SFI is fundamentally a framework for applying two types of strategies: A decomposition strategy that divides a model into smaller sub--models, and an inference strategy that appropriately applies an inference algorithm to each sub--model. SFI uses factors to combine information from solved sub--models to answer queries.

As an example, consider the following model written in SimplePPL, shown in Figs.~\ref{exampleModel} and \ref{LowLevel}. We have three RVs defined in $Q$, $a$, $b$, and $c$. RVs $b$ and $c$ generate a value using the $Q'$ that is generated by $f(a)$, where $outcome^b_T$ refers to the outcome for RV $b$ when $a$ is true, and so forth. Each Chain generates a program $Q'$ for both true and false conditions. With the exception of ``outcome'', the RVs defined by $Q'$ are not directly needed to reason about $a$, $b$, and $c$. That is, all of the RVs defined in $Q'$ except ``outcome'' and $\mathcal{F}_{Q'}$ can be marginalized out of $Q'$. A joint distribution over $\mbox{``outcome''} \cup \mathcal{F}_{Q'}$ is all that is needed to reason at the top--level program $Q$. Since this marginalization is self--contained, any inference algorithm that can compute a joint distribution can be applied to each sub--model (shown in the boxes in Fig.~\ref{LowLevel}). This joint distribution represented as a factor; these factors are then ``rolled up'' and used by another algorithm to answer queries on the program $Q$. This is the core operation of SFI: Given a sub--model, use an algorithm to marginalize away internal variables and return a joint distribution over ``outcome'' and the free variables, and repeat the process until the query is answered. 

%While this model is simple, this process can be repeated for large, complex, highly interconnected models of arbitrary depth.

\subsection{Model Decomposition}

The following discussion of SFI is cast in the context of a SimplePPL program. However, the method is applicable to graphical models in general. 

The key operation of SFI is model decomposition. This operation decomposes a model into semantically meaningful sub--models (i.e., programs) that can be reduced to a joint distribution over relevant variables. First, we define two key concepts: \textit{uses} and \textit{external}. A RV $r$ \textit{uses} a RV $x$ if:
\begin{equation*}
x \in \mathcal{A}_r \wedge P(r | \mathcal{A}_r) \neq P(r | \mathcal{A}_r \setminus x)
\end{equation*}
We denote the set of variables $\mathcal{U}_r$ for a variable $r$ as the set of all variables $r$ uses, either directly or recursively, plus $r$. This definition of uses can be difficult to verify in a program based on the semantics of SimplePPL. However, in our implementation of SimplePPL, we have a syntactic (but stronger) condition for $r$ using $x$ based on $x$ appearing in the expression for $r$, or in any expression for a variable used by $r$. Such a condition is necessary for uses and thus still guarantees SFI's soundness.

We denote that a variable $x$ is \textit{external} to $r$ if:
\begin{equation*}
x \in \mathcal{U}_r \wedge \exists \,\, y \in \mathcal{RV}_Q \setminus \mathcal{U}_r \, | \, x \in \mathcal{U}_y
\end{equation*}
That is, an external variable to $r$ is used in the generative process of a variable that $r$ does not use. We denote the set of variables external to $r$ as $\mathcal{E}^r$.

A decomposition of the model with respect to a RV $d \in \mathcal{RV}_Q$ is an operation that partitions $\mathcal{RV}_Q$ into two disjoint sets of variables, $\mathcal{RV}^d_Q$ and $\mathcal{RV}^{\overline{d}}_Q$. The RV $d$ is called a decomposition point.  We define $\mathcal{RV}^d_Q$ and $\mathcal{RV}^{\overline{d}}_Q$ as:
\begin{equation}
\begin{split}
\mathcal{RV}^{d}_Q & = \mathcal{U}_d - \mathcal{E}^d \\
\mathcal{RV}^{\overline{d}}_Q & = \mathcal{RV}_Q - \mathcal{RV}^d_Q 
\end{split}
\label{decompSets}
\end{equation}
In other words, $\mathcal{RV}^{d}_Q$ is the set of variables exclusively used in the generation of $d$ (i.e., no external uses), and $\mathcal{RV}^{\overline{d}}_Q$ is all remaining variables in $Q$.

As an example, consider the program in Fig.~\ref{exampleModel}. Since $d$ can be any variable, let us choose $outcome^b_T$ as the decomposition point. In this example, $\mathcal{RV}^d_Q$ is the set of variables that $outcome^b_T$ exclusively uses, so it would be $\lbrace outcome^b_T, x1^b_T, y1^b_T, z1^b_T, x2^b_T, y2^b_T, z2^b_T \rbrace$ (all the variables in the left--most box of Fig.~\ref{LowLevel}). $\mathcal{RV}^{\overline{d}}_Q$ would include all other variables in the model. 

\subsection{Factored Representation}

Using factors in SFI has several advantages. First, it provides an interface to communicate the joint distribution of a sub--model to other parts of the model. Second, factors make SFI algorithm--agnostic; any algorithm that can compute a joint distribution and return a factor can be used. For example, sampling algorithms that can post--process a joint distribution into a factor can also be used.

\subsubsection{Factor Creation in SimplePPL}
\label{factorCreation}
Once the variables in $Q$ have been split into two sets via a decomposition point, we convert the decomposition to a factored representation. Each variable $r \in RV_Q$ can be converted to a set of factors $\Psi_r$ that describe the generative semantics of the variable. For variables defined as values or primitives, we create a factor over the support of the variable using the probability distribution defining the variable. For variables defined by $r = Apply(r_1, \dots, r_n, f)$, we create a single factor over $\lbrace r, r_1, \dots, r_n \rbrace$ whose value is $1$ when $r = f(r_1, \dots, r_n)$ and $0$ otherwise. 

Finally, for variables defined by $r = Chain(r_1, f)$, we create a set of factors that represents the joint distribution of $\lbrace r, r_1, outcome_1 \dots, outcome_n \rbrace$. Because representing the joint distribution of $\lbrace r, r_1, outcome_1 \dots, outcome_n \rbrace$ could be prohibitively large, we decompose this joint distribution to keep the sum--product operations tractable. For each value $v$ of $r_1$, we create a factor $\psi_v$ over $\lbrace r, r_1, outcome_v \rbrace$, where $outcome_v$ is the ``outcome'' variable generated from applying $f(v)$. We then generate probabilities for the factor in the following manner:
\begin{itemize}
\vspace{-1mm}
\itemsep-0.5em
\item For each $v' \in r_1$, if $v' \neq v$, then the probability is 1. This is a ``don’t care'' case.
\item For each $v' \in r_1$, $w \in r$, and $x \in outcome_v$, if $v' = v$ and $w = x$, then the probability is 1.
\item Otherwise, the probability is 0.
\end{itemize}
We also create a selector factor over $\lbrace r, r_1 \rbrace$ that selects a factor over $outcome_v$ for the appropriate value of $f(v)$.

\subsubsection{SFI with Factors}

We denote the set of factors created from a program $Q$ as $\Psi$, and each factor $\psi \in \Psi$ is defined over a set of variables $x_{\psi} \in \mathcal{RV}_Q$. Given $\Psi$, the probability distribution of a RV $r$ in the program, $P(r)$, is formulated as:
\begin{equation}
P(r) = \frac{1}{Z}\sum_{x \in \mathcal{RV}_Q \setminus r} \, \prod_{\psi \in \Psi} \psi(x_{\psi})
\label{factorForm}
\end{equation}
where $Z$ is the normalizing constant.

As $d$ divides $\mathcal{RV}_Q$ into two sets, it naturally divides $\Psi$ into two sets, $\Psi_d$ and $\Psi_{\overline{d}}$, as explained in Sec.\~ref{factorCreation}. As such, with $y_{\psi} \in \mathcal{RV}_Q$, we can rewrite Eqn.~\ref{factorForm} as:
\begin{equation}	
P(r) = \frac{1}{Z}\sum_{x \in \mathcal{RV}^{\overline{d}}_Q \setminus r} \, \sum_{y \in \mathcal{RV}^d_Q \setminus r} \, \prod_{\psi \in \Psi_{\overline{d}}} \psi(x_{\psi}) \prod_{\psi \in \Psi_{d}} \psi(y_{\psi})
\label{factoredDecomp}
\end{equation}
Note that even though the variables in $\mathcal{RV}^d_Q$ and $\mathcal{RV}^{\overline{d}}_Q$ are disjoint, the variables used in the sets of factors $\Psi_{d}$ and $\Psi_{\overline{d}}$ are \textit{not} disjoint. From the definition of the sets in Eqn.~\ref{decompSets}, the only variables that are shared between $\Psi_{d}$ and $\Psi_{\overline{d}}$ can be $\mathcal{E}^d$, the set of variables external to $d$, and $d$ itself. As such, we can move the summation over $\mathcal{RV}^d_Q$ in Eqn.~\ref{factoredDecomp} inwards and the summation over $d$ to the outer summation, so that we get:
\begin{equation*}
\begin{split}
P(r) & = \frac{1}{Z} \!\!\!\!  \sum_{x \in \lbrace \mathcal{RV}^{\overline{d}}_Q \cup d \rbrace \setminus r} \, \prod_{\psi \in \Psi_{\overline{d}}} \psi(x_{\psi}) \!\!\!\! \sum_{y \in \lbrace \mathcal{RV}^d_Q \setminus d \rbrace \setminus r} \, \prod_{\psi \in \Psi_{d}} \psi(y_{\psi})\\
 & = \frac{1}{Z} \!\!\!\!  \sum_{x \in \lbrace \mathcal{RV}^{\overline{d}}_Q \cup d \rbrace \setminus r} \, \prod_{\psi \in \Psi_{\overline{d}}} \psi(x_{\psi}) \, \psi^{\mathcal{E}^d}
\end{split}
\label{factoredDecompInwards}
\end{equation*}
where $\psi^{\mathcal{E}^d}$ is a joint factor over $d$ and the external variables defined with respect to $d$. Again looking at Fig.~\ref{exampleModel} as an example, with $outcome^b_T$ as the decomposition point, we perform the summation over $\lbrace x1^b_T, y1^b_T, z1^b_T, x2^b_T, y2^b_T, z2^b_T \rbrace$ and are left with a factor over only $outcome^b_T$. This factor can them be multiplied with the remaining factors in the program and $outcome^b_T$ can be summed out.

In this formulation, a decomposition point $d$ implies a structured process to compute $P(r)$ from a set of factors defined on a model: First, compute a joint distribution with respect to the decomposition point, then compute $P(r)$ using the joint distribution and the remaining factors. Computing the joint distribution over the external variables can be accomplished by any algorithm, as can the computation of $P(r)$ once the joint distribution is computed.

So far, we have only mentioned a single decomposition point in a model. However, multiple decomposition points can be defined on a model. In Fig.~\ref{LowLevel}, for example, there are four natural decomposition points ($outcome^b_T, outcome^b_F, outcome^c_T, outcome^c_F)$ that can be marginalized independently (shown as the boxes in Fig.~\ref{LowLevel}). Eqn.~\ref{factoredDecomp} can be reformulated for multiple points as:
\begin{equation}
P(r) = \frac{1}{Z}\sum_{x \in \lbrace \mathcal{RV}^{\overline{D}}_Q \cup D \rbrace \setminus r} \prod_{\psi \in \Psi_{\overline{D}}} \psi(x_{\psi}) \prod_{k=1}^n \psi^{\mathcal{E}^{d_k}} 
\label{factoredDecompInwardsMult}
\end{equation}
where there are $n$ decomposition points, and $\mathcal{RV}^{\overline{D}}_Q$ is the intersection of $\mathcal{RV}^{\overline{d_k}}_Q , k=1, \dots n$. Decomposition points can be nested inside other decomposition points, allowing inference to proceed in any hierarchical structure implied by the model.

%\subsubsection{Finding Decomposition Points}

In principle, any RV could potentially be a decomposition point. However, we would like to choose a decomposition point $d$ that leads to a small joint factor $\psi^{\mathcal{E}^d}$ and eliminates as many variables in $\mathcal{RV}_Q$ as possible. Chains present a natural decomposition point, which have the benefit of being automatically derived from the program and don't need to be specified by the programmer. When we apply the Chain function $f$ to a parent value $v$, $f(v)$ is a program that defines a sequence of RVs, ending in a definition of a variable named ``outcome''. By the semantics of Chain, only the outcome RV can be used anywhere else in the program. For each Chain defined in $Q$, we create a decomposition point at $outcome_v$ for every value $v$ in the support of $r_1$. This also implies that $\mathcal{E}^d = \mathcal{F}_{Q'}$ for a decomposition point. Thus, we know that the joint factor created at each decomposition point will only be over each ``outcome'' variable and free variables defined in the program $Q'$ generated from $f$.

\section{USING SFI}

\subsection{SFI Operation}
\begin{algorithm}[t]
\caption{Overview of the SFI algorithm\label{SFIAlg}}
\begin{algorithmic}[5]
\Function{Decompose}{program $Q$, variables $\mathcal{E}$, dStrategy $DS$, iStrategy $IS$}
	\State $\Psi \gets \emptyset$
	\For{$c \in Q_{chain}, v \in r_c$}
			\State $Q' \gets f_c(v)$
			\State $\mathcal{E}_{Q'} \gets \mathcal{E}^{Q'.outcome}_{Q'} \cup Q'.outcome$
			\State $\Psi^{\mathcal{E}_{Q'}} \gets DS(Q', \mathcal{E}_{Q'}, DS, IS)$
			\State $\Psi \gets \Psi \cup \Psi^{\mathcal{E}_{Q'}}$	  
	\EndFor
	\State $\psi^{\mathcal{E}} \gets IS(\Psi \cup \Psi_{\overline{D}}, \mathcal{E})$
	\State \Return $\psi^{\mathcal{E}}$
\EndFunction
\Function{SFI}{program $Q$, query $q$, dStrategy $DS$, iStrategy $IS$}	
	\State $\psi_q \gets Decompose(Q, q, DS, IS)$
	\State \Return $Normalize(\psi_q)$
\EndFunction
\end{algorithmic}
%\vspace{-3mm}
\end{algorithm}

Algorithm~\ref{SFIAlg} outlines inference in SFI. To query for the distribution over an RV $q$, a user calls the $SFI$ function with the program $Q$ (written in SimplePPL), $q$, a decomposition strategy $DS$, and an inference strategy $IS$. $DS$ and $IS$ are functions that guide the decomposition and inference of the model, and are explained in more detail below. The $SFI$ function calls the $Decompose$ function, and the resulting factor over $q$ is normalized to compute $P(q)$.

The $Decompose$ function visits each decomposition point in $Q$, applies $DS$ to the sub--model (i.e., program) defined by each point, and marginalizes out the internal variables using $IS$. On lines 3, SFI iterates over all Chains defined in $Q$ and each value $v$ of the parent variable $r_c$. On each iteration, it generates $Q'$, the program created by applying the function $f_c$ to a value $v$ (line 4). Next, it creates the set of relevant variables to program $Q'$ as the external variables in $Q'$ and the ``outcome'' variable (line 5). It then invokes $DS$ on the new program, which returns a set of factors that is added to the current set for $Q$ (lines 6 and 7). Note that a decomposition may also be recursive, as described below.

Once all decomposition points have been visited, the set of factors not generated from a decomposition point ($\Psi_{\overline{D}}$, via the factor generation described in Sec.~\ref{factorCreation}) is added to $\Psi$ and $IS$ is applied which returns a factor $\psi^{\mathcal{E}}$ over the variables in $\mathcal{E}$ (lines 9 and 10). Much of the work of the SFI framework is performed by the decomposition and inference strategies, so we explain these in detail below.

\subsection{Strategies for Decomposition}

A decomposition strategy $DS$ is a method that defines how a program should be decomposed. It is a function that receives a program $Q'$ and set of relevant variables $\mathcal{E}_{Q'}$, and returns a set of factors $\Psi^{\mathcal{E}_{Q'}}$ over \textit{at least} $\mathcal{E}_{Q'}$. The simplest $DS$ is what we call ``raising'': For a point $d$, return the set of factors over all variables defined in $Q'$. This strategy performs no inference, and as a result, all of the factors from $\mathcal{RV}_{Q'}$ are ``rolled up'' to the top--level. If each $d$ is raised, we get a ``flat'' strategy. This is how typical inference works; factors are created for all variables and all non--query variables are marginalized out in a flat operation.  

To take advantage of different inference algorithms, it is clearly beneficial to have a $DS$ that actually reduces the number of variables in the returned factor set $\Psi^{\mathcal{E}_{Q'}}$. As such, we define a recursive strategy as one that will recursively apply the $Decompose$ function until no more decomposition points are found, shown in Alg.~\ref{recursiveDecomp}.

\begin{algorithm}[t]
\caption{A recursive decomposition strategy\label{recursiveDecomp}}
\begin{algorithmic}[5]
\Function{RecursiveDecomposition}{program $Q$, variables $\mathcal{E}$, iStrategy $IS$}
	\State \Return $Decompose(Q, \mathcal{E}, this, IS)$
\EndFunction
\end{algorithmic}
%\vspace{-3mm}
\end{algorithm}

Here, each decomposition point in a model is recursively visited in a depth--first traversal. Once a program is reached with no decompositions, $IS$ is applied to the factors in the program, and a joint distribution over the external variables and outcome is returned, and the process is repeated. This is referred to as hierarchical inference in SFI.

More complex and sophisticated strategies can also be applied. For example, a strategy could decompose only if $\mathcal{E}$ is at most $n$ variables ($n$ would have to be specified at compile time). If the number of external variables is greater than $n$, then the function returns all of the factors defined for the program without running any inference strategy. Otherwise, it calls $Decompose$ again to continue the recursive decomposition.

\subsection{Strategies for Inference}

A strategy for inference applies an inference algorithm to a set of factors defined by program $Q$ and returns a joint distribution over $\mathcal{E}$, the set of external variables in the factors. While SFI uses factors communicate the joint distribution to other programs, there is no restriction that an algorithm operate on factors. As long as the algorithm can ingest factors from other decompositions and output a joint factor over  $\mathcal{E}$, then any algorithm can be used.
	
%\subsubsection{SimplePPL Inference Algorithms}

SimplePPL's implementation of SFI uses factor--based algorithms. There are three algorithms available: Variable elimination (VE)~\cite{koller2009probabilistic}, belief propagation (BP)~\cite{yedidia2003understanding} and Gibbs sampling (GS)~\cite{geman1984stochastic}. VE and BP are standard implementations of these algorithms on factors, and as such we do not provide any details. GS is implemented on a set of factors, but integrating it into SFI is not trivial. Much of the effort is due to the determinism frequently found in PPLs. Our implementation uses automated blocking schemes to ensure proper convergence of the Markov chain. Details on GS can be found in the supplementary material.

\subsubsection{Choosing an Algorithm}

SFI provides the opportunity to develop schemes that dynamically select the best inference algorithm for a decomposition point, serving as the foundation for an automated inference framework. At the application of the inference strategy, there is opportunity to analyze and estimate the complexity of various algorithms applied to the factors, and choose the one with the smallest cost (e.g., speed or accuracy). For example, methods that estimate the running time of various inference algorithms on a model~\cite{lelis2013predicting} can be encoded into an inference strategy, and the algorithm with the lowest estimated running time can be chosen.

We created a simple heuristic to choose an inference algorithm, but yet still demonstrates the potential of the approach. As VE is an exact algorithm, it is always preferred over other algorithms if it is not too expensive, but unfortunately is impractical on most problems. We therefore have a heuristic to use VE on a set of factors. We first compute an elimination order, $O$, to marginalize to the external variables. The cost of eliminating a single variable is the increase in cost between the new factor involving the variable and the existing factors, and the cost of VE is the maximum over all variables, using $O$. If the cost is less than some threshold we use VE, otherwise, BP or GS. 
%We then compute the cost of running VE as 
%\begin{equation}
%Cost = \underset{v \in O}{Max} \, |\, \sum_{v} \prod_{\psi \in \Psi_v} \psi \,| - \sum_{\psi \in \Psi_v} |\, \psi \,| 
%\end{equation}
%where $\Psi_v$ is the set of factors that have variable $v$, $|\psi|$ is the number of entries in the factor, and $v$ traverses the set $O$ in order of the elimination. In other words, the cost of eliminating a single variable is the increase in cost between the new factor involving the variable and the existing factors, and the cost of VE is the maximum over all variables. If the cost is less than some threshold we use VE, otherwise,we use BP or GS. 

To choose between BP and GS, we also use another heuristic. As the degree of determinism in a model strongly correlates with the convergence rate of BP~\cite{ihler2005loopy}, we use the amount of determinism in the model as a choice between using BP or GS. We mark a variable as deterministic if, when using GS, we must create a block of more than one variable. If the fraction of deterministic variables (as compared to all variables) in the model is greater than a threshold, then we invoke BP and otherwise GS. While these strategies are heuristics, they do demonstrate the proof of concept for automated inference, and the results presented in the next section show that they are effective.

\section{EXPERIMENTS}

%\subsection{Models}

We tested SFI using three models. First, we encoded a version of the QMR medical diagnosis model~\cite{jaakkola1999variational} in SimplePPL. Like the standard QMR model, this one is a Bayesian network of causal diseases associated with observable symptoms. However, this model inserts a layer of intermediate diseases between the causal and symptom layer. Thus the intermediate diseases are conditioned on the causal diseases, and the symptoms conditioned on the intermediate diseases. The number of diseases and the number of parents per symptom are varied during testing, and the network is constructed by randomly connecting symptoms to intermediate diseases then subsequently the intermediate diseases to random causal diseases. In each test, a random number of symptoms and causal diseases are observed as evidence.

The second model is a mixed directed--undirected model. The undirected portion of the model is an Ising model~\cite{glauber1963time}, where each Boolean variable $v$ in an $n \times n$ grid has a potential to its four vertical and horizontal neighbors. The prior over each variable, however, is a modeled as a small Bayesian network conditioned on a causal variable $c_n$. Thus this model can be viewed as $n^2$ Bayesian networks that are joined together in a top--level Ising model. The grid size is varied during testing, but for each test a random 20\% of the $c_n$ variables are observed as either true or false.

The last model we used is a simplified version of the Bayesian seismic monitoring system presented in~\cite{arora2013net}. This model is designed to detect and localize seismic events on a simulated two--dimensional representation of earth (with semi--realistic physics). The model consists of a set of monitoring stations at different locations that detect a variety of seismic signals over time. Based on a generative process for both true seismic events and false detections, the model is designed to infer the actual number of seismic events using measurement data (i.e., observations) from each of the detection stations. Continuous variables from the original model were discretized for factored inference; for most tests each distribution was discretized to five bins, but some tests varied the number of bins. We used 10 detection stations with one true event, and varied the total number of false detections from zero to 27. Observation data was generated from a third--party simulation of the seismic generative process. Note that as the number of false detections and discrete bins increases, the model quickly scales up in the number and size of the factors. On some tests, we were unable to attain ground truth since no exact algorithms could complete with available memory.

All of these models are well suited for SFI. First, they contain a significant amount of structure that can be used for decomposition; the diseases and symptoms in the QMR model represent a series of Chain RVs, each Boolean variable in the Ising model is a Chain that uses each $c_n$ as a parent, and each seismic station's observations are Chain RVs. Second, they are fairly common models used in practice and are realistic. All queries for testing were posterior distribution queries over a random subset of diseases (QMR model), Boolean variables (Ising model), or the number of true seismic events.

\subsection{Results}

\subsubsection{QMR and Mixed Model Results}

\begin{figure}[t]
\centering % was .9 columnwidth
\subfigure[]{
\includegraphics[width=0.9\columnwidth, height=37mm]{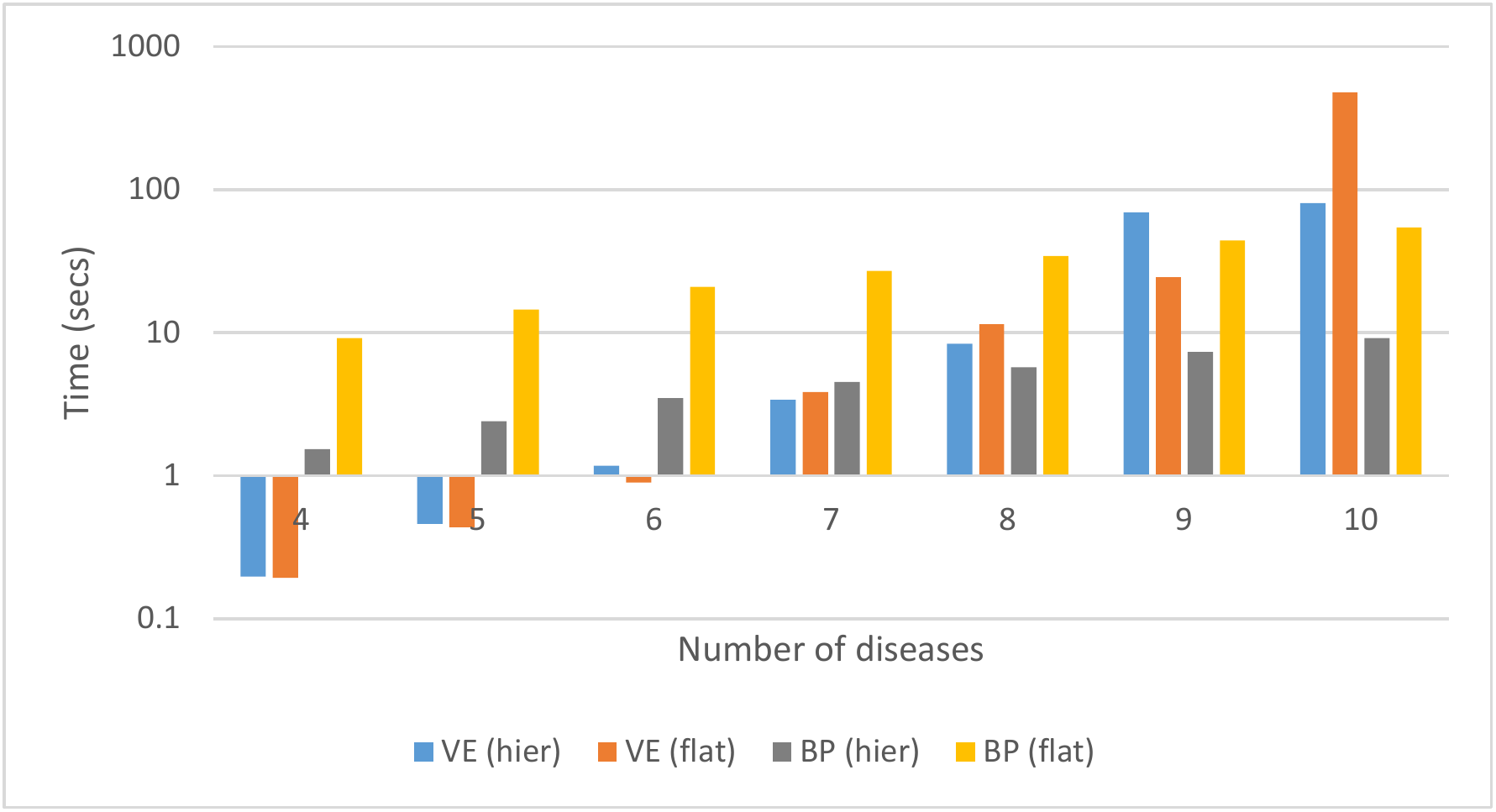}
\label{FlatVsHierTime}
}
\subfigure[]{
\includegraphics[width=0.9\columnwidth, height=37mm]{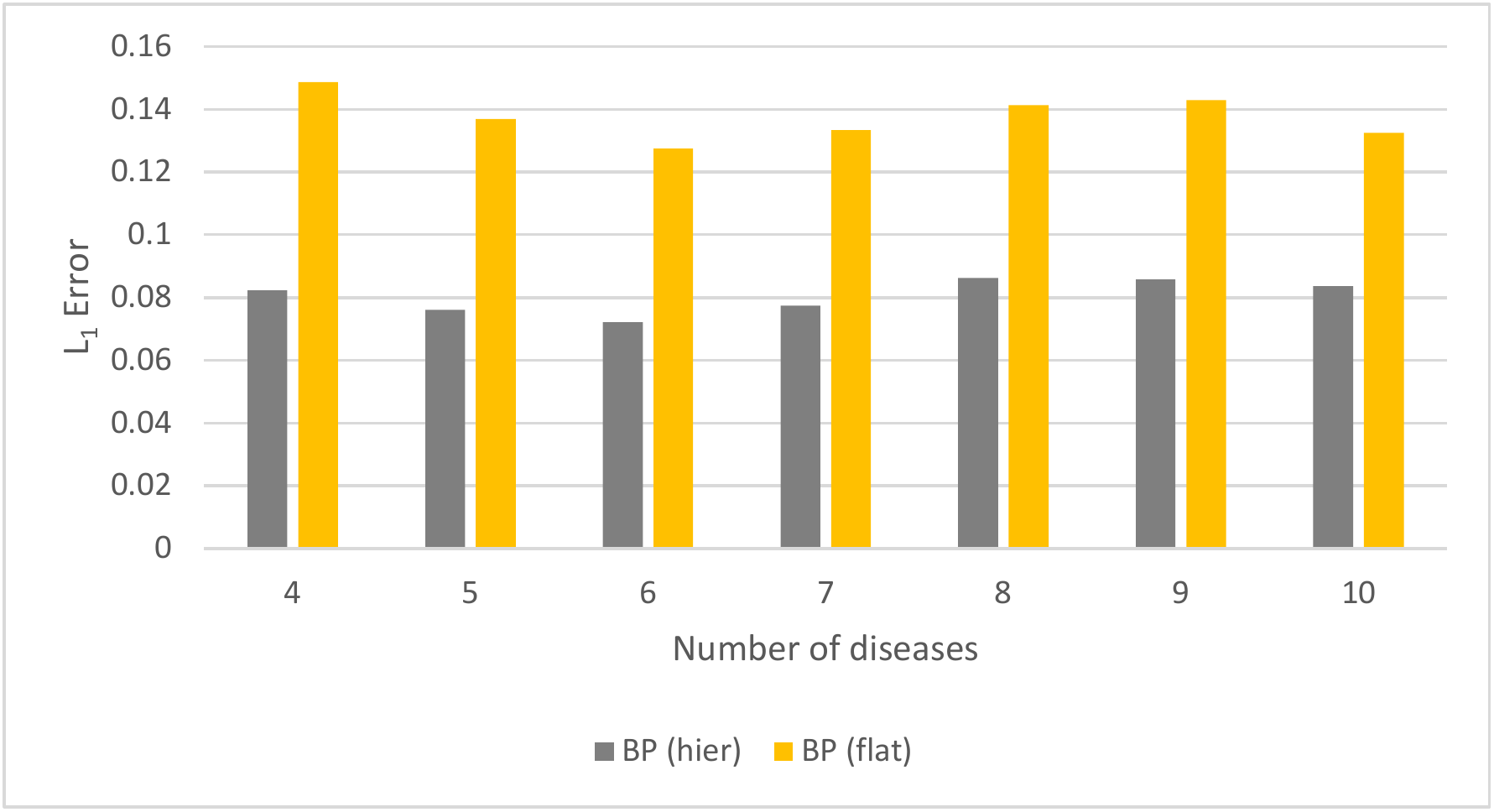}
\label{FlatVsHierAcc}
}
\vspace{-2mm}
\caption{Comparing (a) running time of flat and hierarchical strategies on VE and BP and (b) accuracy of BP as compared to VE.}
\vspace{2mm}
\end{figure}

\begin{figure}[t]
\centering % was .9 columnwidth
\subfigure[]{
\includegraphics[width=0.9\columnwidth, height=37mm]{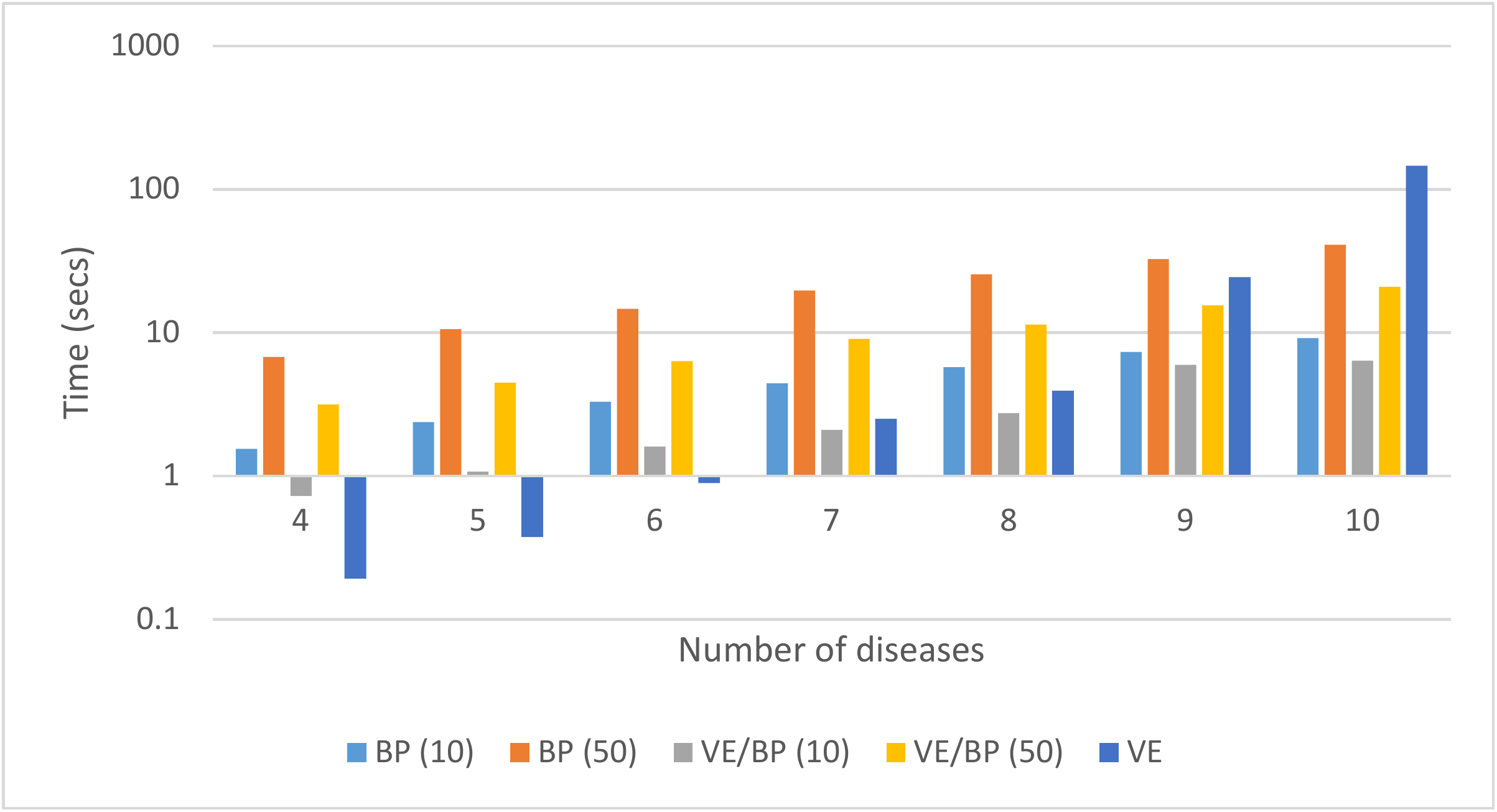}
\label{QMRTime}
}
\subfigure[]{
\includegraphics[width=0.9\columnwidth, height=37mm]{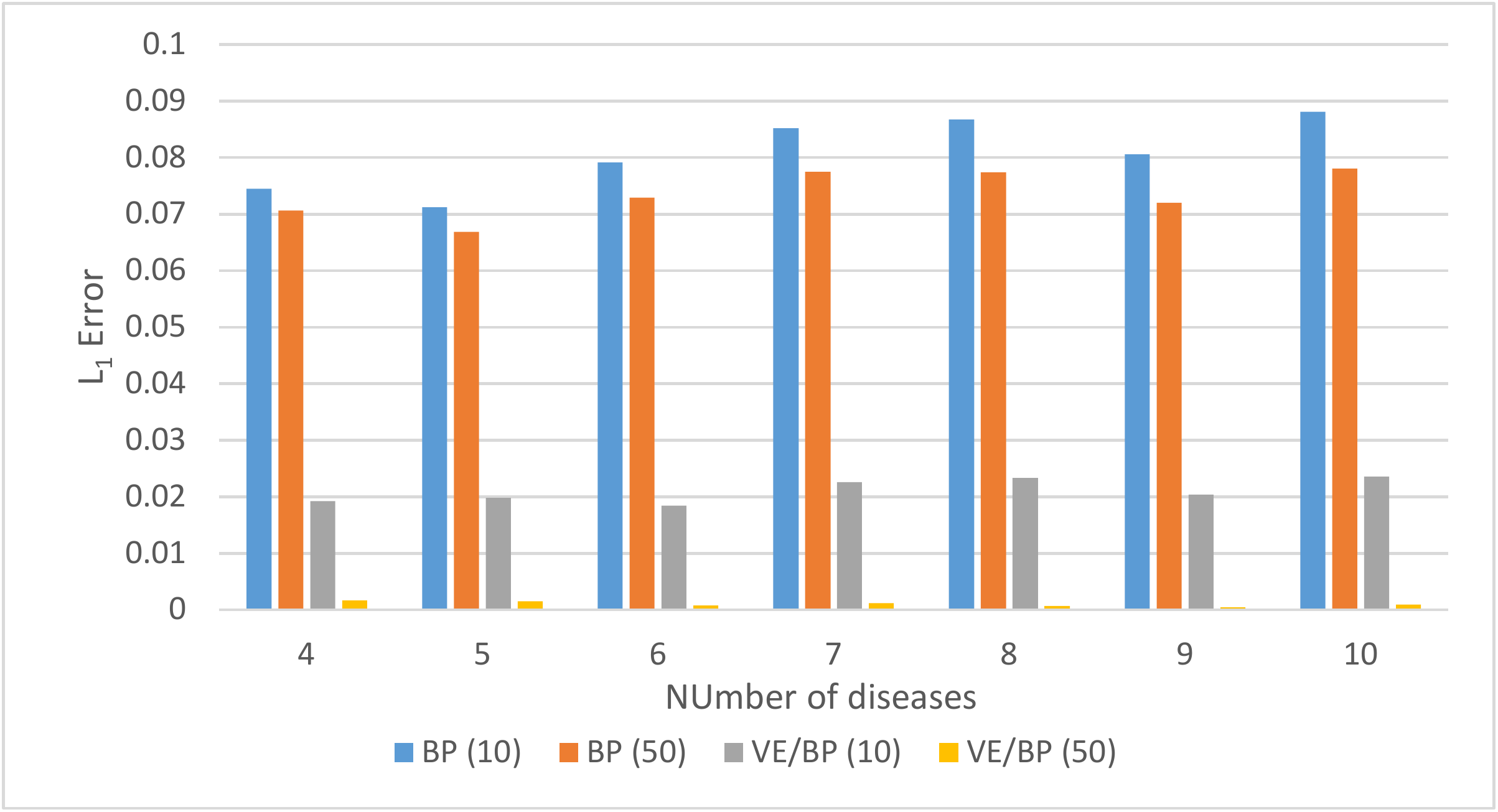}
\label{QMRAcc}
}
\vspace{-2mm}
\caption{Comparing the hybrid inference strategy on the QMR model. (a) Running time, and (b) accuracy as compared to VE.}
\vspace{2mm}
\end{figure}

\begin{figure}[t]
\centering % was .9 columnwidth
\subfigure[]{
\includegraphics[width=0.9\columnwidth, height=37mm]{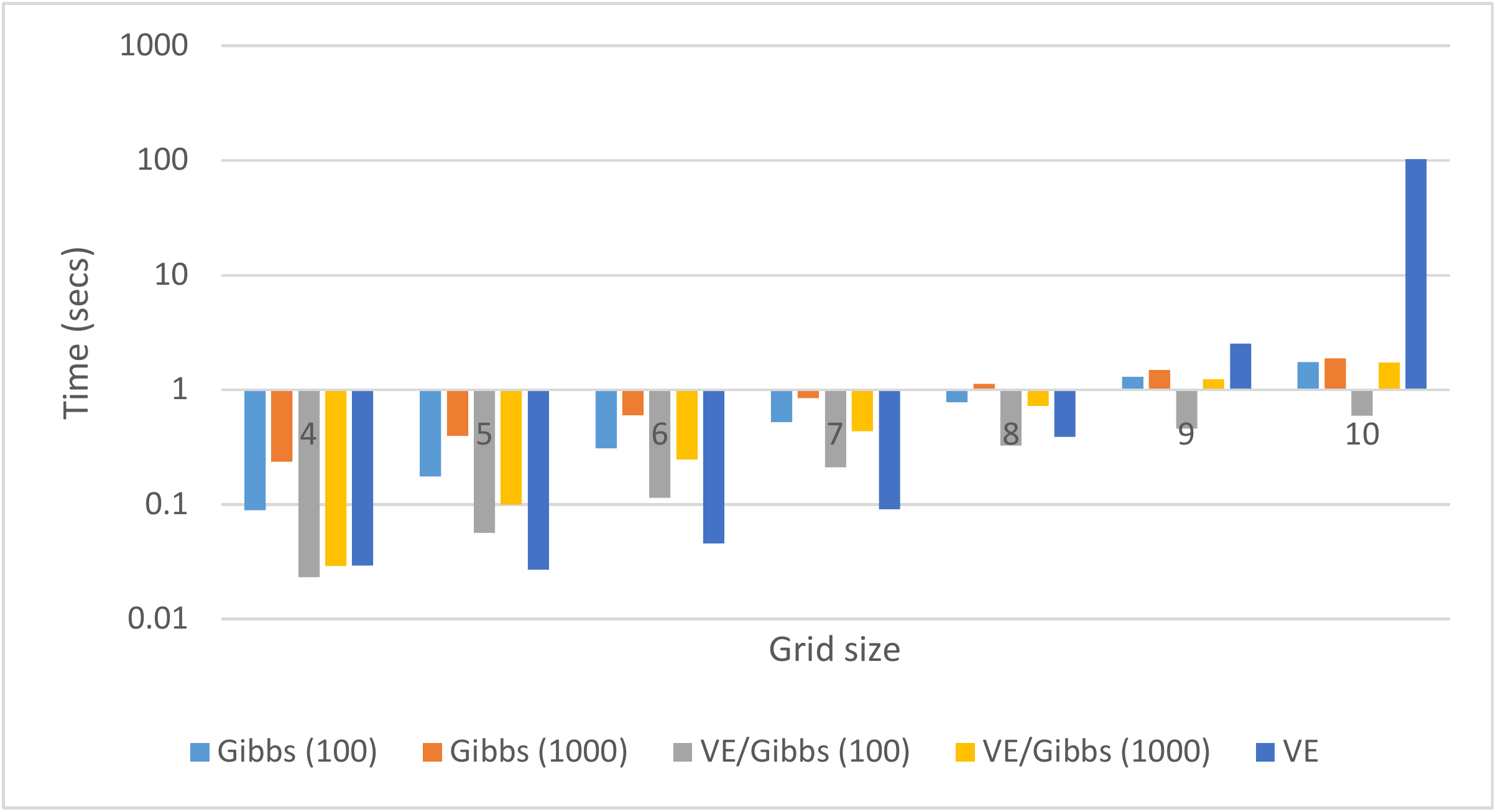}
\label{IsingModelTime}
}
\subfigure[]{
\includegraphics[width=0.9\columnwidth, height=37mm]{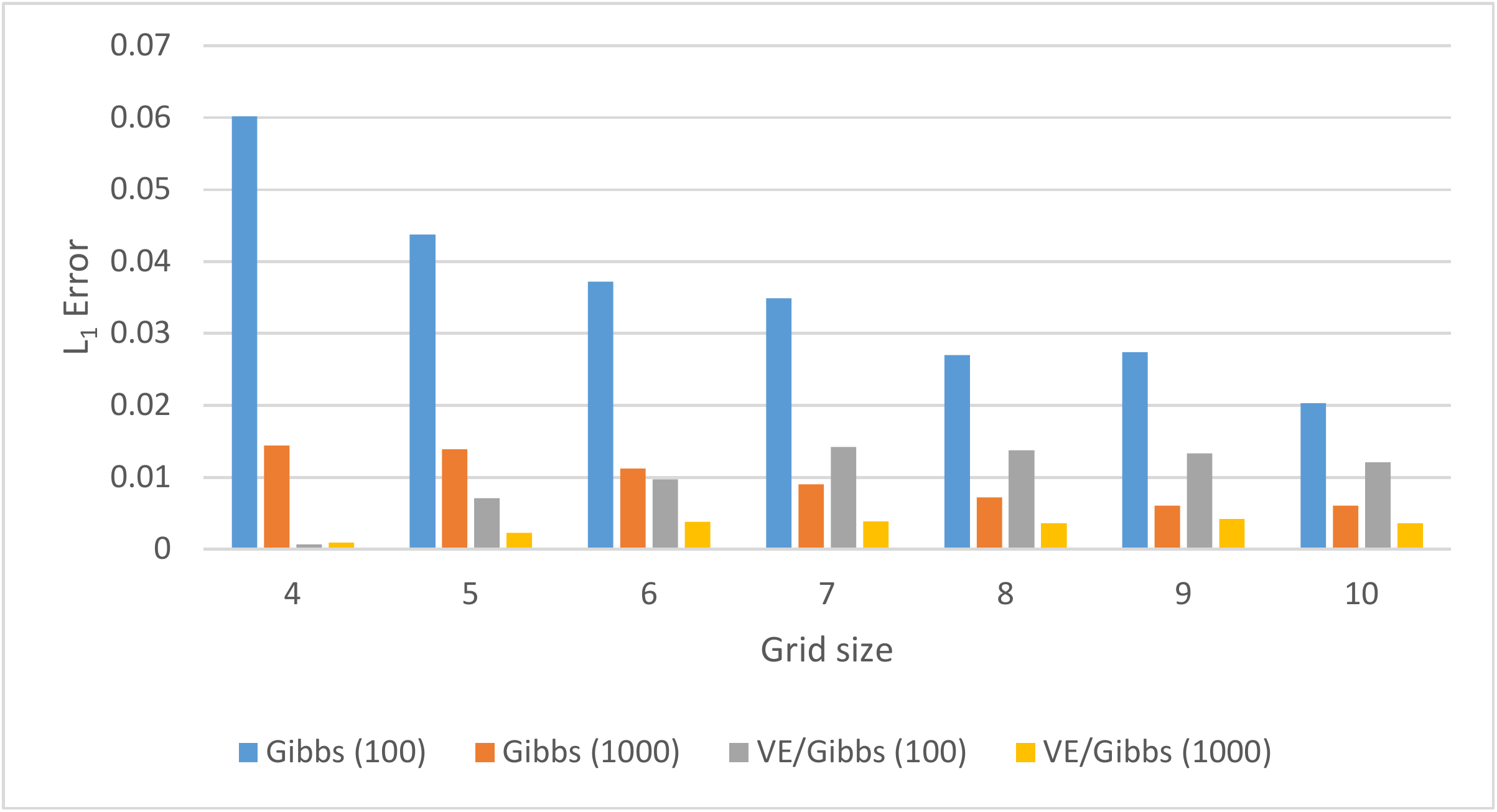}
\label{IsingModelAcc}
}
\vspace{-2mm}
%\caption{Comparing our three algorithm hybrid inference strategy to single inference algorithms on the mixed directed--undirected model. The hierarchical decomposition strategy was used for all tests. (a) Running time of VE, GS (100 and 1000 iterations), and our hybrid strategy that used a combination of VE and GS (100 and 1000 iterations). (b) Accuracy of the GS and the combination methods as compared to VE.}
\caption{Comparing the hybrid inference strategy on the mixed model. (a) Running time, and (b) accuracy as compared to VE.}
\vspace{2mm}
\end{figure}

We first tested how different decomposition strategies affect inference. We used two strategies, flat and hierarchical, on the QMR model using BP and VE. For the hierarchical version of BP, each instance of BP ran for 10 iterations, whereas on the flat version BP was run once for 100 iterations. The results of the running time and accuracy (of BP) are shown in Figs.~\ref{FlatVsHierTime} and \ref{FlatVsHierAcc}, respectively. 

On VE, the results show that the hierarchical strategy generally is faster than the flat one. Mathematically both strategies are performing the same operations. However, the hierarchical strategy imposes a partial elimination order; an elimination order is found for each instance of VE, but the order that the decomposition points are visited is fixed by the strategy. The flat strategy uses a heuristic to find the best elimination order given all the factors in the model. From these results, it appears that the structure imposed by programmer (i.e., by using Chains) finds a better elimination order than the heuristics used to solve this NP--hard problem~\cite{pearl2014probabilistic}. These results are consistent with previous work on structured VE~\cite{koller1997object}.

For BP, the hierarchical strategy consistently runs faster than the flat strategy. This comparison is not exact though as it is hard to determine how many iterations in a flat strategy ``equals'' hierarchical iterations. However, looking at the accuracy of the methods in Fig.~\ref{FlatVsHierAcc}, the hierarchical method is consistently more accurate. Thus, even if the flat BP is run for more iterations to improve its accuracy (assuming it has not converged), it is already dominated in running time and accuracy by the hierarchical method.

Next, we applied our hybrid strategy (with hierarchical decomposition strategies) to determine if we can improve the speed and/or accuracy using multiple algorithms as compared to a single algorithm. The results on the QMR model for running time and accuracy are shown in Figs.~\ref{QMRTime} and \ref{QMRAcc}, respectively. The inference strategy only chose to run VE and BP during inference, so we only compare to those algorithms. 

For running time, VE remains competitive until the number of diseases reaches about 9. Again, comparing BP to the combined VE/BP method directly is difficult. However, combined with the accuracy results in Fig.~\ref{QMRAcc}, we can analyze the relationship between running time and accuracy for all the methods. Both BP--10 and BP--50 have comparable accuracy. The hybrid methods, however, are much more accurate. The hybrid VE/BP--10 method is nearly 4 times more accurate than BP--10/BP--50 and VE/BP--50 approaches nearly zero error. The running time for the hybrid methods are both faster than their ``respective'' BP version (i.e., comparing BP--10 to the hybrid VE/BP--10 test). While VE/BP--50 has a longer running time than the single BP--10 iteration strategy, it is nearly as accurate as VE with significantly less running time.

The results on the mixed model are also shown in Figs.~\ref{IsingModelTime} and \ref{IsingModelAcc}. The strategy only chose to run VE and GS during inference, so we only compare to those algorithms. Similar to the QMR model, VE has the best performance until the model becomes large, at which point the hybrid strategy has the best running time. Looking at Fig.~\ref{IsingModelAcc}, the accuracy of the hybrid methods (GS--100 and GS--1000 iterations) is better than the single GS approaches. Overall, the hybrid approach dominates the GS approach in terms of accuracy and time. However, as more GS iterations are performed, this performance gap will decrease, as the running times of both the single and hybrid approaches are dominated by applying GS to the undirected portion of the model.
 
\subsubsection{Seismic Monitoring Results} 
 
\begin{figure}[t]
\centering % was .9 columnwidth
\includegraphics[width=0.90\columnwidth, height=37mm]{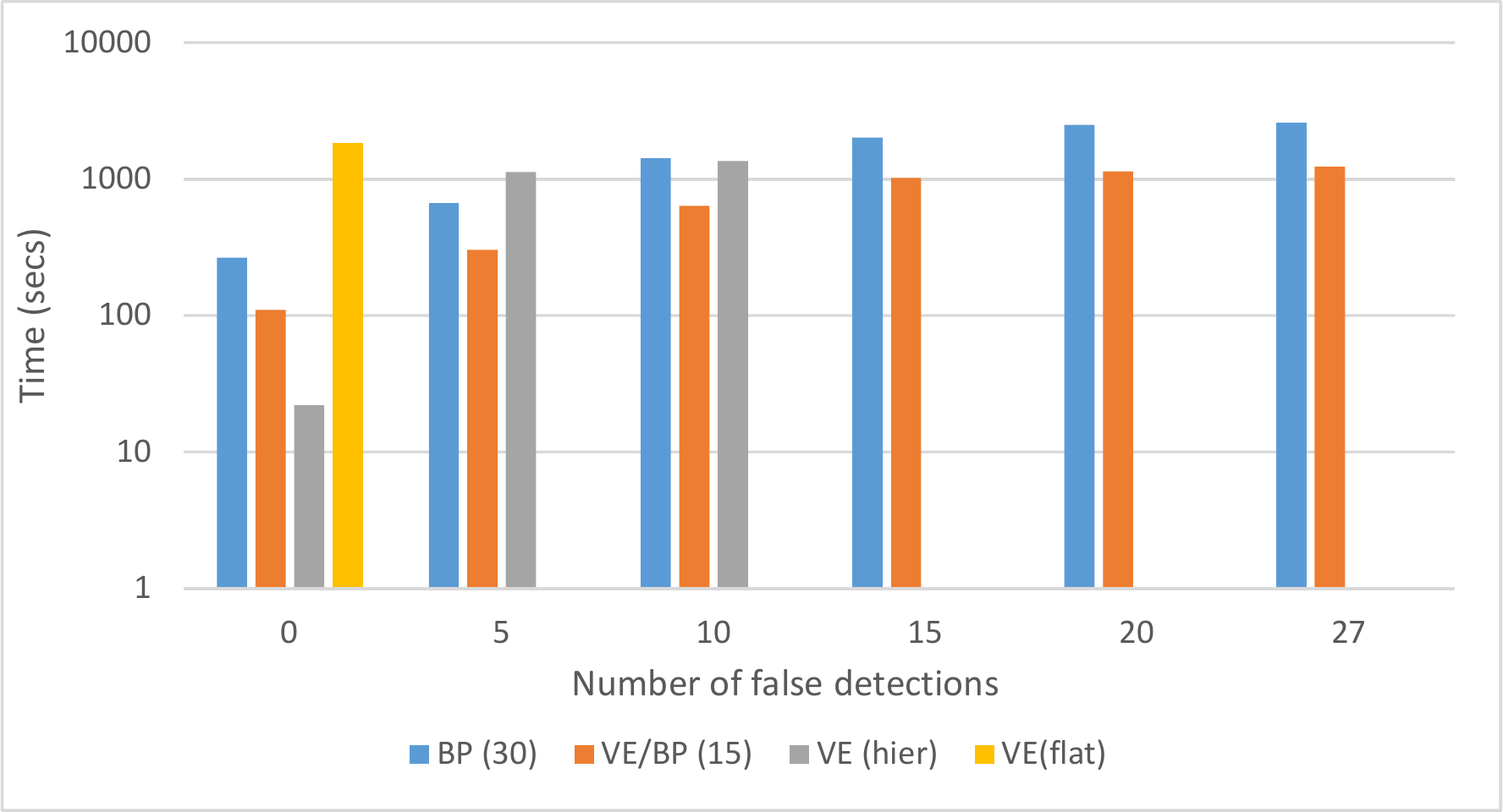}
\vspace{-2mm}
\caption{Comparing the running time as false detections increase. VE did not complete on several tests. \label{noisy_obs_pts5}}
\vspace{1mm}
\end{figure}

\begin{figure}[t]
\centering % was .9 columnwidth
\subfigure[]{
\includegraphics[width=0.9\columnwidth, height=37mm]{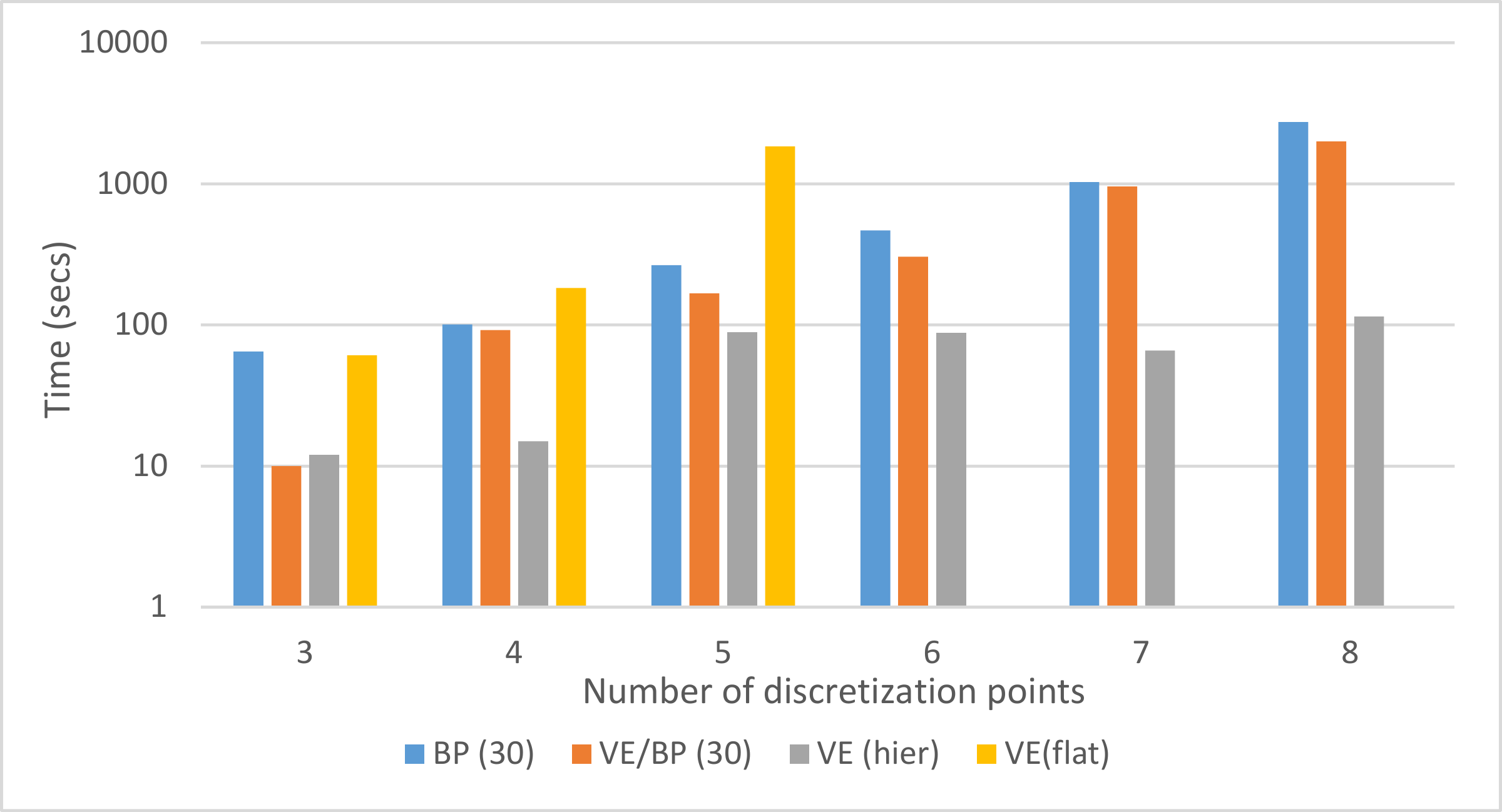}
\label{noNoise_allpts_time}
}
\subfigure[]{
\includegraphics[width=0.9\columnwidth, height=37mm]{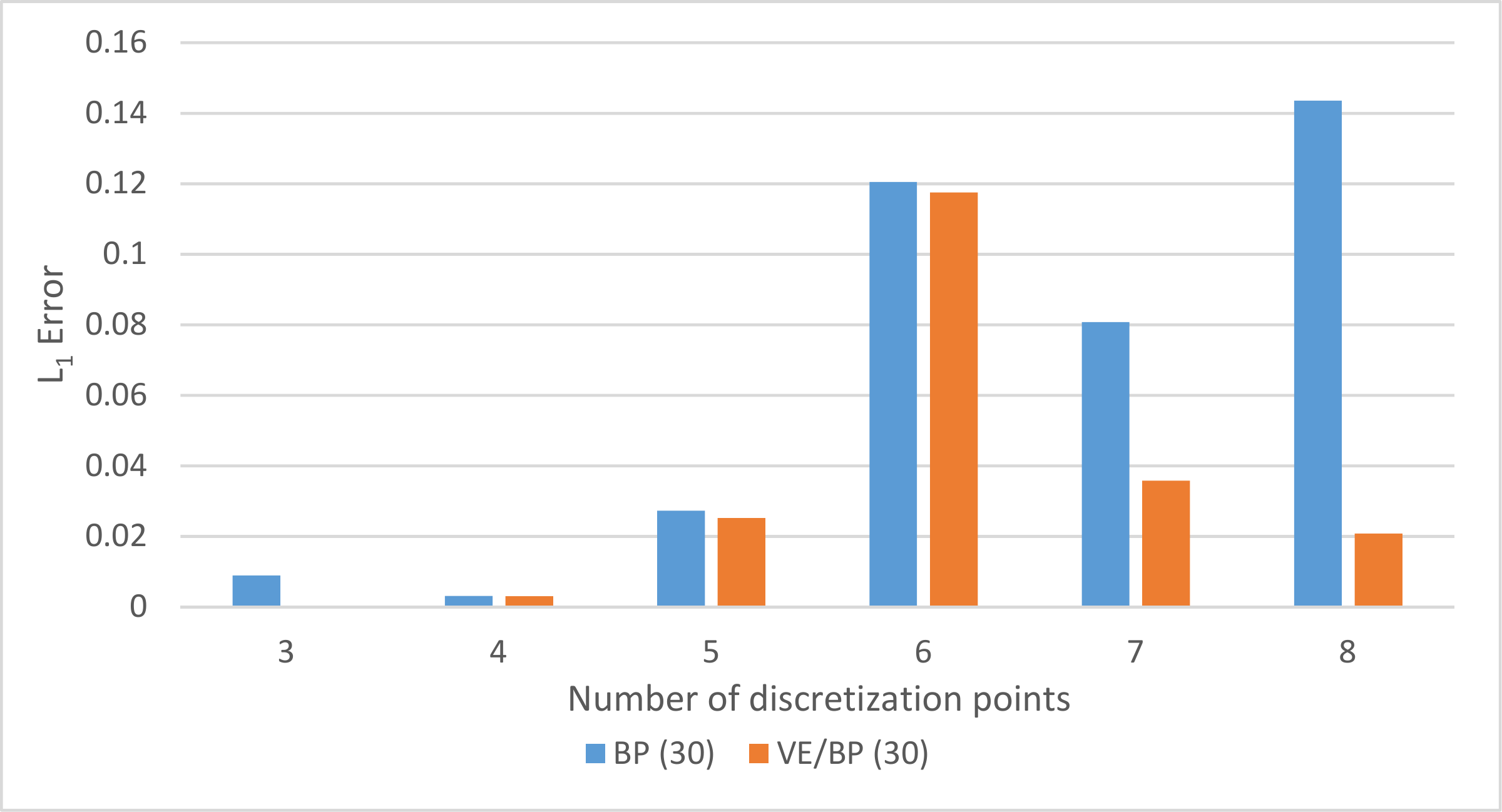}
\label{noNoise_allpts_accuracy}
}
\vspace{-2mm}
\caption{Comparing the (a) running time, and (b) accuracy as compared to VE as the discretization of the model increases. VE did not complete on several tests. \label{noNoise}}
\vspace{2mm}
\end{figure}

On the seismic monitoring model, we ran two experiments where we either varied the number of false detections (Fig.~\ref{noisy_obs_pts5}) or discretization in the models (Fig.~\ref{noNoise}). Fig.~\ref{noisy_obs_pts5} shows the running time of our hybrid VE/BP--15 strategy, hierarchical VE, and flat versions of VE and BP--30 as the number of false detections in the seismic model increases. That is, each test increased the noise in the model, making it harder to detect the true seismic events and significantly increasing the complexity of inference. On the hybrid strategy tests, the SFI framework ran VE for most of the sub--models with the exception of BP on a few sub--models with large tree--widths. Because the two versions of VE ran out of memory on several tests, we are unable to determine ground truth values. The difference in accuracy between VE/BP--15 and BP--30, however, is negligible. We can see that the hybrid strategy has a smaller running time than all the other strategies. As before, it is hard to determine ``equivalence'' between VE/BP--15 and BP--30, but even if 30 iterations is enough to reach convergence, the hybrid strategy still dominates in terms of time.
 
Figs.~\ref{noNoise_allpts_time} and \ref{noNoise_allpts_accuracy} compare strategies as the number of discretization points in the model increases. This has the effect of increasing the size of the factors while keeping the number of factors in the model constant (and assumed no false detections). In Fig.~\ref{noNoise_allpts_time}, hierarchical VE actually has the fastest running time. The hybrid strategy is faster than BP--30 and flat VE, but hierarchical VE clearly dominates the hybrid strategy. Again, this demonstrates that programmer imposed elimination orderings can be much better than heuristic approaches. Clearly, however, this shows that our heuristic to choose an algorithm is not sophisticated enough to understand that VE is preferred on this model. A strategy that uses more than just the increase in factor size to select an algorithm is required in this situation.

Finally, since structured VE completed on all tests, we show the accuracy of the VE/BP--30 and BP--30 methods in Fig.~\ref{noNoise_allpts_accuracy}. As can be seen, the hybrid method that uses a combination of VE and BP is more accurate than flat BP. In addition, as shown in Fig.~\ref{noNoise_allpts_time}, the VE/BP--30 method dominates BP--30 in terms of running time.

\section{CONCLUSION}

In this work, we described a new framework for inference in probabilistic modeling called Structured Factored Inference. Leveraging the capabilities of probabilistic programming, we have shown a semantically sound method to decompose a model into smaller sub--models, and detailed how the application of strategies can guide the inference process. Using simple heuristics to analyze the complexity of sub--models, we demonstrated that SFI can be used to implement a basic automated inference scheme that reasons faster than approximate inference and is nearly as accurate as exact inference methods.

This work serves as a starting point for a more robust automated inference framework, but more analysis is still required. First, there needs to be more theoretical analysis on the criteria for declaring that a sub--model is ``solved.'' That is, how many iterations of GS or BP should be invoked on each sub--model? Most likely answering this question touches upon issues of algorithm convergence, but its impact in the SFI framework is an open research problem.

Second, as shown by the seismic monitoring model, more intelligent algorithm selection methods need to be developed. Recent work provides a starting point for this (\cite{flerova2012preliminary, lelis2013predicting}), but new methods that leverage the analyzability of PP can make the estimation of complexity more accurate~\cite{hur2014slicing}. Finally, new decomposition points would also need to be developed to enable more sophisticated model decomposition. Chain decomposition is effective, but user--defined or object--oriented decomposition points may be more effective at decomposing a model into sub--models that facilitate faster inference. Our hope is that the SFI framework will be the catalyst for future research in these areas.

\subsubsection*{Acknowledgements}

This work was supported by DARPA contract FA8750-14-C-0011.

%\subsubsection*{References}
\bibliographystyle{IEEEtran}
\bibliography{Bibliography}

\end{document}